\documentstyle[12pt,fleqn]{article}
\author{Wolfgang Menzel\\Fachbereich Informatik, Universit\"at Hamburg \\
Vogt-K\"olln-Stra{\ss}e 30, \mbox{D-22527} Hamburg, Germany \\
email: {\tt menzel@informatik.uni-hamburg.de}}
\date{}
\title{Parsing of Spoken Language under Time Constraints
\thanks{This research has been partly funded by the
Federal Ministry for Science and
Technology within the framework of the Verbmobil joint research project
(grant no. BMFT 01 IV 101 A/O)} }

\makeatletter

\newcounter{@sc}
\newcounter{@scp}
\newcounter{@t}
\newlength{\@x}
\newlength{\@xa}
\newlength{\@xb}
\newlength{\@y}
\newlength{\@ya}
\newlength{\@yb}
\newsavebox{\@pt}

\def\bezier#1(#2,#3)(#4,#5)(#6,#7){\c@@sc#1\relax
  \c@@scp\c@@sc \advance\c@@scp\@ne
  \@xb #4\unitlength \advance\@xb -#2\unitlength \multiply\@xb \tw@
  \@xa #6\unitlength \advance\@xa -#2\unitlength
      \advance\@xa -\@xb \divide\@xa\c@@sc
  \@yb #5\unitlength \advance\@yb -#3\unitlength \multiply\@yb \tw@
  \@ya #7\unitlength \advance\@ya -#3\unitlength
      \advance\@ya -\@yb \divide\@ya\c@@sc
  \setbox\@pt\hbox{\vrule height\@halfwidth  depth\@halfwidth
   width\@wholewidth}\c@@t\z@
   \put(#2,#3){\@whilenum{\c@@t<\c@@scp}\do
      {\@x\c@@t\@xa \advance\@x\@xb \divide\@x\c@@sc \multiply\@x\c@@t
       \@y\c@@t\@ya \advance\@y\@yb \divide\@y\c@@sc \multiply\@y\c@@t
       \raise \@y \hbox to \z@{\hskip \@x\unhcopy\@pt\hss}%
       \advance\c@@t\@ne}}}
\makeatother
\newcommand{\cbox}[1]{\parbox{4cm}{\begin{center}#1\end{center}}}

\begin{document}
\maketitle

\begin{abstract}
Spoken language applications in natural dialogue settings place serious
requirements on the choice of processing architecture. Especially under
adverse phonetic and acoustic conditions parsing procedures have to be
developed which do not only analyse the incoming speech in a
time-synchroneous and incremental manner, but which are able to
schedule their resources according to the varying conditions of the
recognition process. Depending on the actual degree of local ambiguity
the parser has to select among the available constraints in order to
narrow down the search space with as little effort as possible.

A parsing approach based on constraint satisfaction techniques is
discussed.  It provides important characteristics of the desired
real-time behaviour and attempts to mimic some of the attention
focussing capabilities of the human speech comprehension mechanism.

\vspace{7mm}
This paper has been presented at the  11th European Conference on
Artificial Intelligence, Amsterdam \cite{menzel:94a}. It appeared also
as Verbmobil-Report No.26.
\end{abstract}

\newpage

\section{INTRODUCTION}

Apart from psycholinguistic evidence about the basic principles of
human speech comprehension incrementality is an obvious requirement for
advanced spoken language systems even due to very practical reasons:

\begin{enumerate}
  \item Natural dialogue settings require an instantaneous response
  capability, which cannot be provided by the usual
``past-the-carriage-return''-type of language processing. The analysis
has to keep pace with the incoming speech data, and even follow-up
activities such as the generation of the desired response have to be
carried out in a concurrent fashion, thus facilitating fluency in
discourse and smooth man-machine-interaction.
  \item Incremental speech comprehension is a fundamental prerequisite
  for the participation in mixed initiative dialogues: The decision to
take the initiative relies crucially on an immediate analysis and
interpretation of partial speech utterances in order to keep the time
delay as short as possible.
  \item Procedures running with minimal time delay are particularly
  advantageous since they satisfy the desire to constrain the memory
capacity for maintaining intermediate results in a natural way.
  \item Speech understanding is a heavily expectation driven process with
expectations derived from a discourse or domain model being most prominent.
Hence, incrementality in speech analysis is essential for an effective
generation of predictions on all levels of language processing.
  \item Incrementality, furthermore, is an inevitable property in
  ambitious applications such as the simultaneous interpretation of
speech.
\end{enumerate}

Under all these conditions a utility function can be assumed which
decreases steadily as the analysis time for the incoming speech signal
grows. Responses to past utterances will neither yield a sensible
contribution in a dialogue nor a useful hypothesis about what the
speaker will probably produce next. Almost always an approximate or
incomplete analysis might be of considerably more benefit than a
perfect but late contribution.

Time-synchroneous and incremental analysis of spoken language requires a
system architecture which at least supports the necessary synchronization
between the speech signal and the processing activities of all system
components involved. This corresponds to the minimum of external control
assumed in highly decentralized, distributed architectures based on the
message passing paradigm \cite{pyka:92}. Here, synchronization is attempted
by controlling the individual time horizon of each component and simply
suppressing the delivery of recognition hypotheses generated with a too big
time delay.

Controlling the temporal behaviour of a system by interrupting its
internal message flow, on the other hand, presumes system components
with the ability to schedule their own workload depending on the time
yet available. This causes no serious problems if time is not
considered to be a critical resource.  However, for most practical
settings this assumption is not justified. Even for the unique speech
comprehension capabilities of the human, time may become a decisive
factor influencing the ``degree of understanding'' considerably. This
is typical for situations which are characterized by the presence of
one or possibly several stress factors, including e.g. fast speech,
nonnative languages, poor articulation and noisy environments.

In such situations there is not an even workload distribution across
the speech signal: In a kind of ``scanning understanding'' the hearer
tries to pick up parts of the input signal, a procedure which makes
heavy use of relevancy estimations.  Each attempt to try to focus too
much attention on a particular part of the input may severely disturb
the ongoing speech perception.

If spoken language systems are desired which are capable of adapting
dynamically to varying time constraints, their components have to cope
with the phenomenon of a steadily shrinking time horizon. To supply
sensible results under such conditions, there must be a predictable
relation between the amount of time spent to solve a task and the
expected quality of the output produced. In particular a tradeoff
between processing time and output quality should be expected.

\begin{figure}[t]
\unitlength 0.5cm
\begin{picture}(30,5.5)

\put(4,1){\vector(1,0){8}}
\put(4,1){\vector(0,1){4}}
\put(4,4){\line(1,0){1}}
\bezier{200}(5,4)(7,4)(8,2.5)
\bezier{200}(8,2.5)(9,1)(10,1)
\put(3.4,4.4){$u$}
\put(11.6,0.3){$t$}

\put(17,1){\vector(1,0){8}}
\put(17,1){\vector(0,1){4}}
\bezier{400}(17,1)(19,4)(24,4)
\put(16.4,4.4){$q$}
\put(24.6,0.3){$t$}
\end{picture}
\caption{Utility function and performance profile for a hypothetical anytime
procedure}
\end{figure}
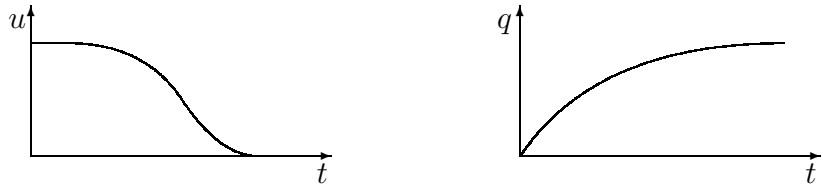

Procedures which show the desired monotonic growth of output quality
have been termed anytime modules ({\sc Boddy and Dean}
\cite{boddy:dean:89}, \cite{boddy:dean:94}, {\sc Russel and
Zilberstein} \cite{russell:zilberstein:91}). Their role for the
development of spoken language systems has first been noted by {\sc
Wahlster} \cite{wahlster:93}.  The most prominent feature of an anytime
module is the existence of a {\em quality measure} (e.g. certainty,
accuracy or specificity). Its (probabilistic) variation over time is
described by a performance profile. {\sc Russell and Zilberstein}
distinguish between two cases of anytime behaviour:

\begin{enumerate}
  \item {\em Interruptible algorithms}, where interruptions may occur
  without previous warning. The component will always be able to
deliver a solution of the quality specified by the performance
profile.
  \item {\em Contract algorithms}, which will only yield a sensible
  result of the specified output quality level if they are supplied
with a previously determined time interval. Otherwise they will not be
able to produce any useful result.
\end{enumerate}

In both of these cases the performance profile for the task at hand is
expected to be known in advance. However, this requirement is a rather
restrictive one which for certain types of combinatorical algorithms
simply cannot be presumed.

\section{ANYTIME PARSING}

Traditional parsing algorithms do not meet the anytime condition at
all. For instance, a depth-first analysis spends all but the very last
part of its processing time with the inspection of useless blind
alleys. Breadth-first on the other hand seems to be better suited from
the anytime point of view, but in fact it provides a monotonic growth
of the {\em completeness} of individual parses instead of continuously
improving a quality parameter of an overall input description. If
interrupted before finishing at least a single complete parse, a chart
will contain either a set of not yet verified and incomplete parse
trees (top down mode) or a set of competing and possibly contradictory
partial analysis results (bottom up mode). In general, no knowledge
will be available on how these fragments may be combined in order to
form a useful parsing result.  More seriously, there is no suitable
quality measure at hand with which the improvement of parsing results
can be described \cite{goerz:kesseler:94}, not to mention a predictable
performance profile.

In fact, there have been proposed alternative parsing schemata which
much better fit into anytime demands than the usual chart parsing
approach. One example is the attempt to parse sentences by tree-to-tree
transductions, which already has been used in the framework of the
machine translation system ARIANE-78 \cite{boitet:etal:82} more than a
decade ago. The sentence to be parsed is provided as a completely flat
tree where all the terminal leaves are immediately dominated by the
topmost node. Parsing takes place by successively replacing partial
trees by more structured ones aiming at a description of constituency
structure in the usual sense. On the one hand, this approach offered
the possibility to develop the modules for analysis, structural
transfer, and generation by means of a single uniform formalism
(ROBRA).  On the other hand, it provided a quite natural fail-soft
feature as an inherent property of the basic processing mechanism: If
the parser fails to find a good parse by applying its tree-to-tree
transduction rules, it simply passes the (partially) unmodified tree to
the transfer stage. In such cases a word-by-word translation with a
considerably worse quality is produced.

\begin{figure}[t] \hspace*{5mm}
\unitlength0.5cm
\begin{picture}(25,10)

\put(-4,1){\cbox{Tom}}
\put(-2,1){\cbox{reads}}
\put(0,1){\cbox{the}}
\put(2,1){\cbox{letter}}
\put(0,2){\line(0,1){0.5}}
\put(2,2){\line(0,1){0.5}}
\put(4,2){\line(0,1){0.5}}
\put(6,2){\line(0,1){0.5}}
\put(-4,3){\cbox{n}}
\put(-2,3){\cbox{v}}
\put(0,3.1){\cbox{det}}
\put(2,3){\cbox{n}}
\put(0.25,4){\line(2,1){2}}
\put(2.25,4){\line(1,2){0.5}}
\put(3.75,4){\line(-1,2){0.5}}
\put(5.75,4){\line(-2,1){2}}
\put(-1,5.4){\cbox{s}}

\put(5,1){\cbox{Tom}}
\put(7,1){\cbox{reads}}
\put(9,1){\cbox{the}}
\put(11,1){\cbox{letter}}
\put(9,2){\line(0,1){0.5}}
\put(11,2){\line(0,1){0.5}}
\put(13,2){\line(0,1){0.5}}
\put(15,2){\line(0,1){0.5}}
\put(5,3){\cbox{n}}
\put(7,3){\cbox{v}}
\put(9,3.1){\cbox{det}}
\put(11,3){\cbox{n}}
\put(13,4){\line(2,1){0.8}}
\put(15,4){\line(-2,1){0.8}}
\put(10,4.7){\cbox{np}}
\put(9,4){\line(0,1){0.5}}
\put(5,4.7){\cbox{np}}
\put(9.5,5.5){\line(3,1){1.5}}
\put(11,4){\line(0,1){1}}
\put(11,5){\line(1,2){0.5}}
\put(13.5,5.5){\line(-3,1){1.5}}
\put(7.5,6.4){\cbox{s}}

\put(14,1){\cbox{Tom}}
\put(16,1){\cbox{reads}}
\put(18,1){\cbox{the}}
\put(20,1){\cbox{letter}}
\put(18,2){\line(0,1){0.5}}
\put(20,2){\line(0,1){0.5}}
\put(22,2){\line(0,1){0.5}}
\put(24,2){\line(0,1){0.5}}
\put(14,3){\cbox{n}}
\put(16,3){\cbox{v}}
\put(18,3.1){\cbox{det}}
\put(20,3){\cbox{n}}
\put(22,4){\line(2,1){0.8}}
\put(24,4){\line(-2,1){0.8}}
\put(19,4.7){\cbox{np}}
\put(22.5,5.5){\line(-2,1){1}}
\put(20,4){\line(0,1){1.5}}
\put(20,5.5){\line(2,1){1}}
\put(17.25,6.5){\cbox{vp}}
\put(20.7,7.5){\line(-2,1){1}}
\put(18,4){\line(0,1){2.2}}
\put(14,6.5){\cbox{np}}
\put(18.25,7.5){\line(2,1){1}}
\put(15.5,8.5){\cbox{s}}

\end{picture}
\caption{Trace of a (deterministic) tree-to-tree transduction}
\end{figure}
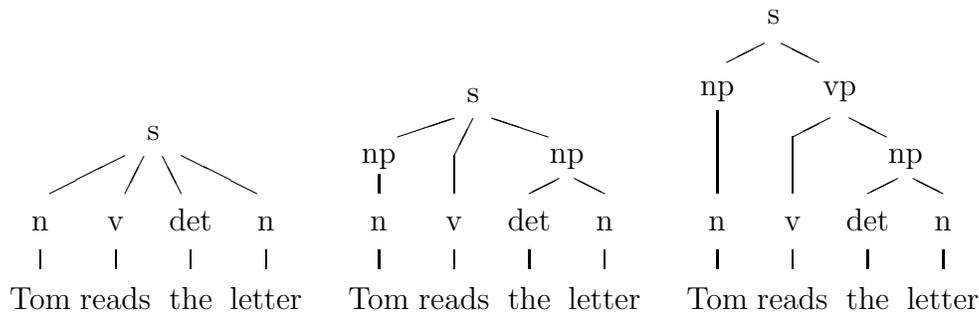

Assuming an almost monotonic improvement of the translation results by
successively applying additional transduction rules, the degree of
structural complexity of a tree can be used as a rough quality measure
in the sense of the anytime condition: Whenever a tree-to-tree
transducer happens to be interrupted it will be able to supply one or
several descriptions of the input sentence. These descriptions always
cover the input completely but they are more or less well structured,
depending on the amount of time spent. This indeed corresponds to a
kind of anytime behaviour in the desired sense: The more time is
available, the higher the structural complexity of the parsing trees
and - hopefully - the better the translation quality will be.

Unfortunately, structural quality makes sense as a quality measure only
in the specific domain of machine translation. If e.g. a semantic
representation, like a logical form, is desired, partially structured
trees will be of little value. Moreover, tree-to-tree transformation
suffers from quite the same disadvantage as a normal chart parsing
procedure does: There is no predictable dependency which could be used
to relate (at least in a probabilistic manner) the expected output
quality to the allocated parsing time.

First of all, this difficulty results from a specific property of the
parsing problem. Natural language parsing is characterized by a rather
unstructured kind of search space which is individually created during
the parsing process. In contrast to other common search problems (c.f.
VITERBI-search in the area of speech recognition) neither the depth nor
the breadth of the space can be estimated prior to the parsing itself.
Hence, techniques like iterative deepening are appropriate to influence
the output quality of a tree-to-tree transduction parser, but certainly
do not make its performance profile more predictable. Although there is
an individual (namely instance specific) monotonic profile, it cannot
be generalized over classes of possible inputs.

This situation suggests a notion of anytime behaviour independently of
the predictability of a module's performance profile. Therefore, a new
distinction is introduced between algorithms with a strong anytime
behaviour and others with a weaker one. A component satisfies the {\em
strong anytime property} if for a certain quality parameter a general
monotonic performance profile exists and is known prior to the
computation itself. {\em Weak anytime algorithms} have a monotonic
performance profile as well but since it is instance specific it cannot
be estimated in advance and allows no prediction of the quality level
to be expected.

According to this distinction parsing by tree-to-tree transduction
turns out to be an interruptable weak anytime algorithm. It can be
finished arbitrarily and the later an interrupt is requested the better
the results can be expected to be.

The notion of a contract module can be given a sensible interpretation
for weak anytime algorithms as well. Again, a contract algorithm does
require a minimal amount of time in order to be able to produce useful
results. Then weak anytime behaviour can be observed, if the module is
apt to optimize its internal processing with the goal of achieving the
best possible output quality in the time interval allocated. This
requires a kind of scheduling mechanism which carries out a dynamic
means-ends analysis with respect to the results achieved so far and the
time yet available.

Neither traditional chart parsing nor the tree-transduction approach
seem to comply with the conditions for a weak anytime contract
algorithm. In general, it will even be difficult to decide how to
continue best when only trying to finish a particular analysis in due
time. The scheduling mechanism would have to find good guesses for a
two-dimensional decision problem

\begin{itemize}
  \item Which sequence of inactive edges in the chart (or which tree
  fragment in the transduction approach) looks most promising for
applying the next rule to it?
  \item Which rule in the grammar should be selected to continue with a
partial solution?
\end{itemize}

Obviously the necessary heuristics are not easily available. Even after
having applied a rule successfully there is almost no possibility to
conclude that this might have been a contribution towards the attempted
final result.  Not surprisingly, almost all of today's parsing systems
still rely on purely combinatorical algorithms, whose time behaviour is
difficult to predict. Under these circumstances there is reason to
assume that parsing algorithms of the weak anytime contract type should
be based on radically different computational principles.

\section{PARSING AS CONSTRAINT SATISFACTION}

Constraint propagation represents a certain exception among the
computational paradigms for combinatorical problem solving, since it
meets the requirements of graceful degradation under time constraints
already due to its very fundamental principles. Within a search space
defined by the assignment of (finite) domains to a finite number of
variables $V_i=\{x|x \in Dom(V_i)\}$ a solution is desired, which
simultaneously satisfies all the conditions from a set of constraints
$C$. Constraints can be thought of as forming a network through which
value restrictions are propagated. At any time in the course of the
computation the network contains all the solutions which are still
consistent with respect to the already applied constraints. In
particular the network will always contain - among others - all the
globally consistent solutions to the constraint satisfaction problem.

Constraint satisfaction has first been applied to the structural
disambiguation of natural language by {\sc Maruyama}
\cite{maruyama:90a}. Local constraints on admissible utterance
structures are defined in the framework of a dependency grammar, where
word forms $w_i \in W$ are modified by others according to certain
dependency relations $l_i \in L$.

A possible modification to a node in a dependency tree is a pair
consisting of a dominating node and a corresponding arc label. These
pairs are taken as possible values of the constraint satisfaction
problem $V_i= \{ p | p \in W \times L\}$. Hence, the current state of
the analysis is described by all the remaining relations by which a
word form can modify another one.

A constraint $c \in C$ then is a relation defined over value
assignments for an arbitrary subset of variables $c \subset V_m \times
. . . \times V_n$.  In order to produce a manageable implementation,
constraints should be restricted to local (i.e. unary or binary) ones.
If $pos(x)$ is defined to denote the position index of a node, $mod(x)$
its modifiee, $lab(x)$ the modifying relation of a node and $cat(x)$
the category of an input word form attached to a node\footnote{Only a
single role identifier per word form is considered. The approach can
easily be generalized to a multidimensional dominance relation. \\ The
treatment of position indices differs slightly from that of
\cite{maruyama:90a} in order to later allow the generalization to
non-sequential input descriptions.}  the unary constraint

\[cat(x)=\mbox{D} \to (lab(x)=\mbox{DET} \] \vspace{-7mm}
\[\;\;\; \wedge cat(mod(x))=\mbox{N} \wedge pos(x)<pos(mod(x))) \]

describes the fact that a determiner (D) can modify a noun (N) on its
right hand side with the dependency relation DET \cite{maruyama:90a}.
Most unary constraints do not restrict the set of possible value
assignments in the usual sense but instead {\em license} a particular
initial state of the network.

The mutual compatibility of value assignments then can be encoded by
binary constraints, as for instance the verb second condition of German
main clauses :

\[(mod(x)=mod(y) \wedge lab(mod(x))=\mbox{V} \wedge pos(x)<pos(mod(x)) \]
\vspace{-7mm}
\[\;\;\; \wedge \; pos(y)<pos(mod(y)) ) \to x=y \]
\hspace*{10mm} {\it ``two words which modify the main verb cannot be
placed both} \\
\hspace*{12mm} {\it left of the verb''} \\

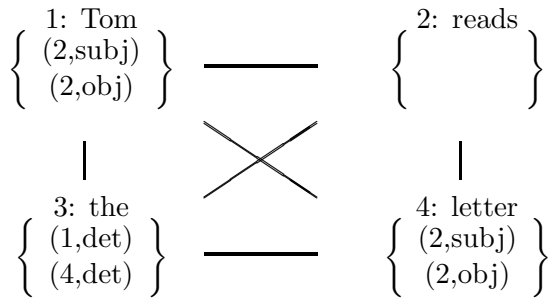
\begin{figure}[t]
\unitlength 0.5cm
\begin{picture}(30,7)
\put(4,6){\cbox{\small 1: Tom \\ $\left\{ \begin{array}{c}
\mbox{\small(2,subj)} \\ \mbox{\small(2,obj)} \end{array} \right\}$}}
\put(14,6){\cbox{\small 2: reads \\ $\left\{ \begin{array}{c}
\;\;\;\;\;\;\;\;\;\;\; \\ \; \end{array} \right\}$}}
\put(4,1){\cbox{\small 3: the \\ $\left\{ \begin{array}{c} \mbox{\small
(1,det)} \\ \mbox{\small (4,det)} \end{array} \right\}$}}
\put(14,1){\cbox{\small 4: letter \\ $\left\{ \begin{array}{c} \mbox{\small
(2,subj)} \\ \mbox{\small (2,obj)} \end{array} \right\}$}}
\put(11,1){\line(1,0){3}}
\put(11,1.03){\line(1,0){3}}
\put(11,6){\line(1,0){3}}
\put(11,6.03){\line(1,0){3}}
\put(7.8,3){\line(0,1){1}}
\put(7.83,3){\line(0,1){1}}
\put(17.8,3){\line(0,1){1}}
\put(17.83,3){\line(0,1){1}}
\put(11,2.5){\line(3,2){3}}
\put(11,2.54){\line(3,2){3}}
\put(11,4.5){\line(3,-2){3}}
\put(11,4.54){\line(3,-2){3}}
\end{picture}
\caption{Content of a constraint network prior to the application of
linear precedence constraints}
\end{figure}

Another binary constraint is the projectivity condition usually assumed
to hold for dependency grammars \cite{maruyama:90a}

\[pos(mod(x))<pos(y)<pos(x) \] \vspace{-7mm}
\[\;\;\; \to pos(mod(x)) \le pos(mod(y)) \le pos(x). \]

By applying constraints of this type to the sets of possible
modifications at the network nodes certain value combinations can be
excluded and the search space is reduced successively. If sufficient
constraints are available eventually a state will be reached where each
node modifies exactly another one (except the topmost node, of course)
and a unique description of the input string has been established.

In contrast to the usual chart parsing approach, parsing by constraint
satisfaction no longer is a procedure which monotonically adds new
partial results but instead monotonically restricts the space of valid
structuring possibilities.

Besides the fact that constraint satisfaction procedures can be
parallelized without serious difficulties, they offer yet another
important advantage:  By simply analysing some formal parameters (e.g.
the size of the value sets at different nodes of the constraint
network) it becomes possible to evaluate the current state of
computation as well as the recent progress. Additionally, a few simple
but rather effective heuristics are available to select a place in the
network where constraint application will probably yield the most
effective reduction of the search space.

Under the perspective of the anytime condition this advantage is  of
crucial importance. For the first time an internal workload scheduling
of the parsing procedure becomes possible. The analysis can be made to
concentrate on those places in the network where disambiguation is most
urgent. Scheduling can be improved further, if a particular ordering on
the set of constraints is assumed which gives a rough estimation of the
restrictive power of a constraint. Again a two-dimensional decision
problem is given. The procedure tries to find the optimal sequence of
constraint applications which allows to determine the global state of
consistency with a minimum of computational effort. In contrast to the
unification grammar tradition, available constraints are not applied at
once, but the parser will decide selectively where to apply which kind
of constraint considering the current state of analysis.

For the purpose of an interactive machine translation system {\sc
Maru\-yama} \cite{maruyama:90a} proposes a kernel grammar approach. It
starts with a minimal but fairly general set of constraints and adds
more specific restrictions only if this becomes necessary to solve
remaining ambiguities.

Constraints can be syntactic as well as semantic or domain specific
ones and no fixed order of constraint application is defined.
Therefore, domain specific constraints which usually are much more
restrictive than those from a general grammar can be taken into
consideration as soon as all of their application conditions hold. If
e.g. disambiguation succeeds using only domain specific knowledge,
syntactic constraints will never be invoked and certain types of
ungrammaticality are accepted without additional effort.  Furthermore,
constraints are not necessarily static ones. Additional constraints can
be requested on demand from other modelling components (e.g. dynamic
domain, discourse or user models) and therefore are particularly
interesting for the design of interactive system structures (c.f.
\cite{pyka:92}).

\section{QUALITY OPTIMIZATION}

Parsing by constraint propagation always departs from a structural
description of maximal ambiguity and aims at successively reducing the
number of different readings for the input as far as possible.
Therefore the degree of remaining ambiguity seems appropriate as a
measure to valuate the progress of computation. It can be used to
schedule the sequence of constraint applications in such a way that the
amount of time required to reach a unique description will be
minimized. On the other hand, it is quite doubtful whether the degree
of remaining ambiguity can be taken as a useful criterion for output
quality {\em per se}. In the long run only a largely disambiguated
description can serve as a sensible basis for further processing.

Even in case of time stress or lack of general constraints this goal
can be reached by the application of heuristic or even brute force
methods.  Heuristic constraints may be based on rules of thumb which
e.g. reduce the search space to the most frequent cases. For instance a
subject-first heuristics for German might be stated in the following
way

\[cat(x)=\mbox{V} \wedge mod(y)=x \wedge lab(y)=\mbox{SUBJ}
\to pos(y)<pos(x) \]

This constraint is a rather restrictive one and will ultimately exclude
all the other constituents of the sentence from being topicalized.

Another well known heuristics is the minimal attachment rule which
prefers shorter dependency relations over longer ones. Heuristics of
this kind no longer restrict the consistency of individual input
descriptions, but instead define {\em preferences} based on a
comparison of different readings.  Therefore it will be difficult to
express them as logical constraints in the usual way. However,
preferences can be expressed as (nonmonotonic) rules which directly
manipulate the search space by eliminating less preferred modification
possibilities from the value sets.

\[mod(x)=y \wedge mod(x)=z \wedge y \neq z \wedge pos(y) < pos(z) < pos(x)\]
\vspace{-6mm}
\[\;\;\; \Rightarrow \mbox{DELETE}(mod(x)=y) \]
\hspace*{10mm} {\it ``For a node $x$ which modifies two others
($y$ and $z$) simulta-} \\
\hspace*{12mm} {\it neously the modification of the more distant
node is suppressed.''}
\vspace{5mm}

The same result can be obtained by a dynamic constraint which puts
a time dependent upper limit on the distance between modifier and
modifiee

\[mod(x)=y \wedge mod(x)=z \wedge y \neq z \wedge pos(y) <
pos(z) < pos(x)\]
\vspace{-6mm}
\[\;\;\; \to pos(x) < pos(mod(x))+n \]

where $n$ should be directly proportional to the remaining time $T$. In
a very similar fashion other dynamic distance heuristics (e.g. the
attachment of a determiner) may be described as well.

Heuristic constraints can be roughly ordered according to their
(estimated) reliability. Then, the selection problem becomes a
three-dimensional one:

\begin{enumerate}
  \item different nodes in the constraint network have different
  degrees of ambiguity,
  \item different constraints are expected to have a different
  potential for ambiguity reduction, and
  \item different constraints have a different degree of reliability.
\end{enumerate}

In order to guide the processing in a nearby optimal manner, all three
criteria have to be weighted against the utility profile of the parsing
module. As long as time pressure is negligible, a general solution to
the parsing problem is attempted using more restrictive constraints
first. Only growing time pressure combined with a comparatively high
degree of ambiguity might justify the activation of heuristic
constraints in order to speed up the analysis.

A quality measure for a weak anytime contract module should be defined
in a such a way that it properly reflects the basic bias between the
remaining ambiguity and the reliability of the constraints used. Let
$a(t)$ denote the remaining ambiguity normalized by the initial one and
$r(t)$ the mean degree of reliability during parsing which both are
defined for the interval $\langle 0, 1 \rangle$.  A sensible quality
measure might then be defined as

\begin{figure}[t] \hspace*{3cm}
\unitlength 0.5cm
\begin{picture}(15,10)
\put(5,3){\vector(-3,-2){3}}
\put(5,3){\vector(0,1){6}}
\put(5,3){\vector(1,0){6}}
\put(2.9,1.6){\line(1,0){5}}
\put(7.9,1.6){\line(3,2){2.1}}
\put(10,3){\line(0,1){5}}
\put(4.8,8){\line(1,0){0.4}}
\bezier{200}(5,3)(9,4)(10,8)
\bezier{200}(7.9,1.6)(9.8,3.2)(10,8)
\bezier{200}(2.9,1.6)(9.2,2)(10,8)
\put(3.4,8.8){$q(t)$}
\put(4,7.7){1}
\put(2.5,0.7){$a(t)$}
\put(2.1,1.5){1}
\put(11,3.2){$r(t)$}
\put(9.8,2.1){1}
\end{picture}
\caption{Quality measure for disambiguation using heuristic constraints}
\end{figure}
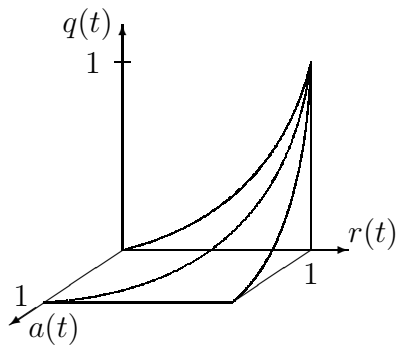

\[q(t)=\frac{e^{r(t) [1 - a(t)]} - 1}{e - 1} \]

Hence, high quality results require both, a low degree of remaining
ambiguity and a high degree of reliability. Low quality on the other
hand is characterized by high ambiguity and/or low reliability.

The introduction of heuristic constraints into the processing provides
the advantage of being able to generate useful output even in the
presence of rather strong time constraints. Although a quality measure
can hardly be determined for individual input utterances, the procedure
will nevertheless exhibit the desired weak anytime behaviour in a
general probabilistic sense.

On the other hand, there is a number of problems which have to be dealt
with:

\begin{enumerate}
  \item Heuristic constraints will almost certainly be in contradiction
  with constraints applied earlier or elsewhere in the network.
Therefore they should be introduced carefully and in a strictly local
manner just to solve a very particular disambiguation problem. At any
rate global inconsistencies have to be avoided, since they prevent the
analysis from producing complete descriptions.
  \item The application of heuristic constraints is in principle
  irreversible.  Heuristics should therefore be considered only if a
unique description cannot be reached otherwise.
  \item The use of heuristic constraints impairs the additive
  continuation behaviour, a fundamental property of customary anytime
modules. Let $\Delta q(\langle t_1,t_2 \rangle)$ denote the (measured)
quality increase of a contract module during time interval $\langle
t_1, t_2 \rangle$, where $t_1 \neq 0$ has to be interpreted as a
continuation of work after an initial contract time $t_1$. Given the
strong anytime condition, the following equation will always hold

\[ \Delta q(\langle 0, t_1 \rangle ) + \Delta q(\langle t_1,
t_2 \rangle) = \Delta q(\langle 0,t_2 \rangle) \]

However, this condition does not need to be valid any longer if
heuristic constraints have been used in order to observe the initial
contract time $t_1$.  In general, only a reduced quality increase
should be expected for the time period after continuation and for a
sufficiently large $t_2 - t_1$ even

\[ \Delta q'(\langle 0,t_1 \rangle ) + \Delta q'(\langle t_1, t_2
\rangle ) < \Delta q(\langle 0,t_2 \rangle). \]

will be observed. The reduced increase is caused by irreversible
restrictions due to the use of heuristic constraints during the initial
time period $t_1$ which prevent the analysis from reaching a nearby
optimal quality level again.  In the definition of a quality measure
above, this is considered by the impossibility to reach a maximum level
of quality with a mean reliability less than one.
\end{enumerate}

\section{PARSING OF SPOKEN LANGUAGE}

Spoken language parsing has to cope with at least two kinds of phenomena

\begin{itemize}
  \item the inherent uncertainty of partial recognition results and
  \item the missing of reliable phrasal boundaries in the speech
signal.
\end{itemize}

Usually word lattices are used to avoid unreliable decisions in the
presence of uncertainty at the interface between speech recognition and
language processing. Each word hypothesis is effectively time stamped
by its starting and ending points and possibly supplemented by a
confidence estimation.

As a first consequence all linear ordering constraints introduced in
the examples above need to be generalized from position indices to
relations over time intervals: Positional ordering is replaced by
interval precedence, identity constraints by a time overlap condition.
Furthermore, additional constraints can be invoked to rule out abnormal
dependency structures.  Because constraints are always defined on
inconsistent modification possibilities, restrictions on overlapping
nodes have to be stated in terms of modification relations as well:

\[mod(a)=b \wedge mod(c)=d \to \neg (overlap(a,b) \]
\vspace{-7mm}
\[\;\;\; \vee overlap(a,c) \vee overlap(a,d) \vee \ldots \vee overlap(c,d)) \]
\hspace*{10mm} {\it ``Whenever two modifying relations are considered,
the nodes} \\
\hspace*{12mm} {\it involved must not overlap each other.''}
\footnote{Three particularly important special cases can be obtained if
$b=c$, $b=d$ or $(a=c \wedge b=d)$ will be inserted into the constraint
and $overlap(x,x)=\mbox{FALSE}$ is assumed.} \vspace{5mm}

This constraint prevents all nodes with no more than two dependency
relations in between from overlapping each other. Obviously, this
condition is not strong enough because modification is a transitive
relation and the transitive closure would have to be taken into
consideration. To improve the effectiveness of simultaneity constraints
a (partial) linearization of dependency trees will become necessary. It
allows to generalize the reasoning about overlap conditions from single
word forms to partial trees but bears a serious risk of bloating the
search space. Therefore it can be tried only in cases of almost
unambiguous dependency relations.

Exactly this turns out to be the main difficulty of lattice parsing by
constraint satisfaction: Important and efficient constraints can only
be applied if the search space has already been narrowed down to a
certain degree. One could try to approach this problem by resorting to
extremely restrictive (usually domain specific) constraints and
extending singular ``islands of certainty'' successively. Considering
the enormous variety of modifying possibilities within a word lattice
it will, however, remain the clear exception that a modifying relation
can be ruled out with absolute certainty by means of compatibility
conditions alone.

Here, probabilistic measures are suitable to supply additional
information.  This includes

\begin{itemize}
  \item bigram-statistics of word form sequences,
  \item probability estimations for dominance possibilities and
  \item confidence values for the word forms involved.
\end{itemize}

Instead of using a Boolean decison, the compatibility of two
modification links is then described as a fuzzy value. Heuristic
constraints can be devised to put awards on the preferred dominance
possibilities and penalties on the unlikely ones. Modification links
with a low valuation can be excluded under growing time pressure.

The second important condition for spoken language parsing is the
missing of appropriate phrase boundaries in the speech signal. Parsing
therefore becomes a time synchroneous procedure and the number of nodes
in the constraint network --- although finite at every time point
during the processing --- will no longer be known in advance. The
network is continuously extended by incoming word hypotheses and
outputs all nodes leaving the time horizon given by the contract time.
For each newly created node all modification links licensed by unary
constraints are established and constraint propagation tries to reduce
the number of readings towards a unique interpretation where each word
modifies exactly another one. In order to meet the anytime condition a
unique interpretation for each node should have been achieved before
the allocated time interval is exceeded. Again the analysis is guided
by the remaining degree of ambiguity and the actual time delay.

\section{CONCLUSIONS}

Constraint satisfaction techniques are well suited to provide natural
language parsing with a weak type of anytime behaviour at least in the
case of deterministic input. Since the paradigm facilitates explicit
reasoning about the available and the required means for
disambiguation, it enables the parsing procedure to dynamically adapt
to external time constraints which are typical for spoken language
applications. Knowledge from very different sources (syntax, semantics,
discourse, domain, user status, \ldots ) can interact in a coordinated
way. The approach is not restricted to static (i.e. universally valid)
constraints and allows to introduce dynamic (i.e. only locally valid)
knowledge even in areas with a traditionally prevalent static point of
view. Thus, for instance, it becomes possible to model the increased
probability of certain topicalized constructions in the context of very
specific surface indicators, like a negation or a possessive pronoun.

A rather simple local measure exists which allows to determine those
parts of the network where disambiguation is most urgent. Combined with an
estimation of the restrictive power of constraints this allows
for a dynamic resource scheduling. The most efficient constraints are
applied first and thus a kind of optimal time behaviour is achieved.
Probabilistic measures can be included without difficulties.

There is some reason to assume that the approach can be extended in
principle to the treatment of spoken language. First of all, this
requires to provide the means for parsing in dynamically extending
lattices. For the time being the main problem rests with the lack of a
tradition in writing linguistic knowledge as local constraints. No
conclusive judgement about the feasibility of the approach can be found
unless at least a nontrivial fragment of a natural language has been
described and tested in order to clarify the most essential questions

\begin{itemize}
  \item To which degree language specific knowledge can be expressed by
means of strictly local constraints on modification possibilities?
  \item How effective is the restrictive power of a grammar compared to
  the typical recognition uncertainty embodied in a word lattice?
  \item Can the restrictive power of a single constraint be estimated
  in a reliable way to allow an effective scheduling procedure being
  devised?
\end{itemize}

Unfortunately, the predominant trend in contemporary computational
linguistics is driving into just the opposite direction. Within the
framework of unification-based grammars increasingly complex
constraints are used to describe combinability conditions and structure
building operations by means of a single uniform formalism. It is just
this complexity which makes it difficult for a system designer

\begin{itemize}
  \item to estimate a constraint's restrictive power,
  \item to determine those parts of a constraint which are sufficient
  to solve a particular disambiguation problem,
  \item to determine those parts of a constraint which - on demand -
  can be replaced by stronger (heuristic) ones,
  \item to find reliable heuristics for determining those partial
  results which look most promising with respect to a final solution
and
  \item to devise algorithms for an incremental analysis where partial
  constraints are applied as soon as possible and partial results are
extended later when additional data comes in.
\end{itemize}

Since on the other hand the merits of the unification-based approach
for writing concise and hence comprehensible grammars should not be
debated, one of the most interesting questions will be, whether it
becomes possible to (semi-) automatically derive local constraints as
needed for constraint satisfaction from the complex feature structures
used in unification based approaches to natural language processing.

\bibliographystyle{plain}

\end{document}